\documentclass[runningheads]{llncs}
\usepackage[T1]{fontenc}
\usepackage{graphicx}
\usepackage{hyperref}
\usepackage{color}
\usepackage{xcolor}
\usepackage{booktabs}
\usepackage{amssymb}
\usepackage{amsmath}
\usepackage{multirow}
\usepackage{marvosym}
\begin{document}
\title{Knowledge Boosting: Rethinking Medical Contrastive Vision-Language Pre-Training}
\titlerunning{KoBo: Knowledge-Boosting Contrastive Vision-Language Pre-training}

\author{Xiaofei Chen\inst{1}, Yuting He\inst{1}, Cheng Xue\inst{1}, Rongjun Ge\inst{2}, Shuo Li\inst{3} \and \\ Guanyu Yang\inst{1,4,5}\textsuperscript{(\Letter)}}
\authorrunning{Chen et al.}
\institute{Key Laboratory of New Generation Artificial Intelligence Technology and Its Interdisciplinary Applications (Southeast University), Ministry of Education\\ \email{yang.list@seu.edu.cn} \and Nanjing University of Aeronautics and Astronautics \and Dept. of Biomedical Engineering, Case Western Reserve University, OH, USA \and Joint International Research Laboratory of Medical Information Processing, Southeast University, Nanjing 210096, China \and Centre de Recherche en Information Biomédicale Sino-Français (CRIBs)}
\maketitle
\begin{abstract}
The foundation models based on pre-training technology have significantly advanced artificial intelligence from theoretical to practical applications. These models have facilitated the feasibility of computer-aided diagnosis for widespread use. Medical contrastive vision-language pre-training, which does not require human annotations, is an effective approach for guiding representation learning using description information in diagnostic reports. However, the effectiveness of pre-training is limited by the large-scale semantic overlap and shifting problems in medical field. To address these issues, we propose the \textbf{K}n\textbf{o}wledge-\textbf{B}o\textbf{o}sting Contrastive Vision-Language Pre-training framework (KoBo), which integrates clinical knowledge into the learning of vision-language semantic consistency. The framework uses an unbiased, open-set sample-wise knowledge representation to measure negative sample noise and supplement the correspondence between vision-language mutual information and clinical knowledge. Extensive experiments validate the effect of our framework on eight tasks including classification, segmentation, retrieval, and semantic relatedness, achieving comparable or better performance with the zero-shot or few-shot settings. Our code is open on \url{https://github.com/ChenXiaoFei-CS/KoBo}.
\end{abstract}
\begin{figure}
	\centering
	\includegraphics[width=0.99\textwidth]{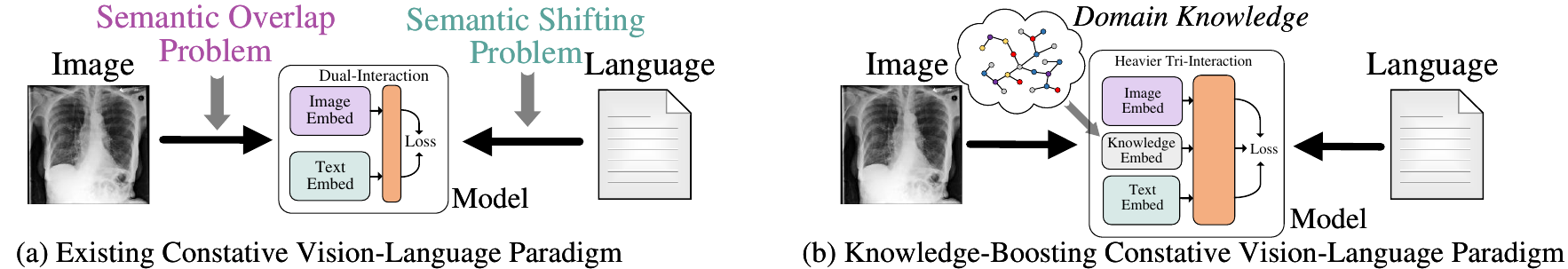}
	\caption{Our knowledge boosting innovates the paradigm of medical vision-language contrastive learning, inspired by two problems in the existing architecture.}
	\label{paradigm}
\end{figure}
\begin{figure}
	\centering
	\includegraphics[width=\textwidth]{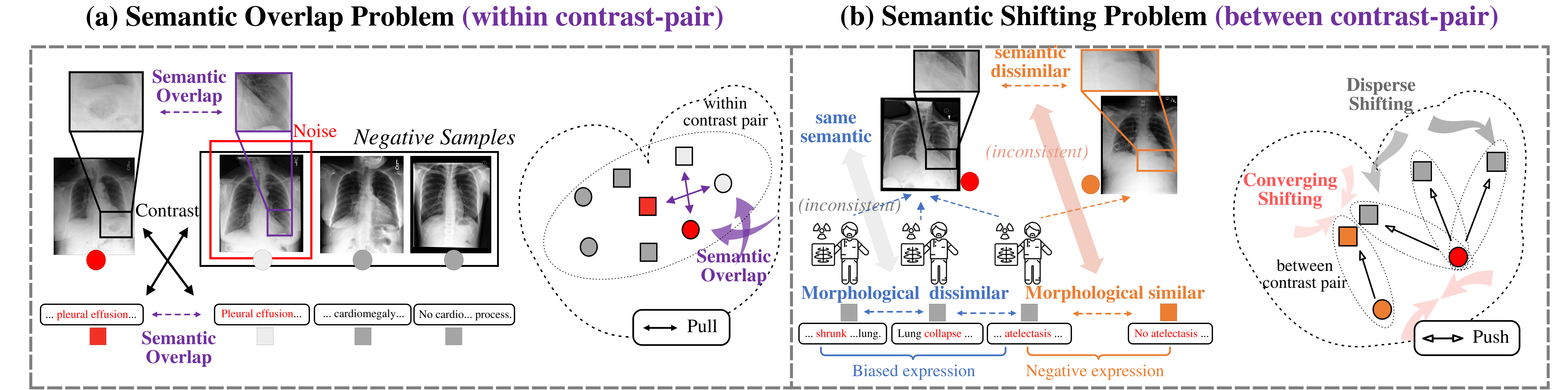}
	\caption{Two key challenges in medical contrastive vision-language pre-training: \textbf{(a)} Semantic overlap exists between negative samples, falsely pulling apart samples with similar semantics. \textbf{(b)} Biased expression and negative expression of radiologists cause the inconsistency of semantics and text morphology between sample pairs, causing disperse and converging semantic shifting.}
	\label{introduction}
\end{figure}

\section{Introduction}
Foundation models have become a significant milestone in artificial intelligence, from theoretical research to practical applications~\cite{foundation_models}, like world-impacting large language model ChatGPT~\cite{nature_chatgpt} and art-history-defining large generative model
DALL-E~\cite{DALL_E}. In medical image analysis, foundation models are showing promising future, and pre-training technologies~\cite{acmmm2022MedVLP,miccai2022M3AE,he2023geometric}, as the cornerstone of foundation models, facilitated feasibility of computer-aided diagnosis for widespread use.
	
Medical contrastive vision-language pre-training~\cite{miccai2022_RGCT,miccai2022_MMSEG,nips2022_MGCA,emnlp2022_MedCLIP,iccv2021Gloria} has shown great superiority in medical image analysis, because it utilizes easy-accessible expert interpretation from reports to precisely guide the understanding of image semantics. Therefore, contrastive vision-language pre-training will break through the bottleneck of time-consuming and expensive expert annotation~\cite{pmlr2022ConVIRT} and difficulty in learning fine-grained clinical features with pure-image self-supervised methods~\cite{miccai2019_model_gensis}. It will improve data efficiency, and achieve comparable or better performance when transferred with the zero-shot or few-shot setting, demonstrating the potential of promoting the ecology of medical artificial intelligence.
	
However, semantic overlap and semantic shifting are two significant challenges in medical vision-language contrastive learning (Fig.\ref{introduction}). \textbf{(a) Semantic Overlap Problem:} There is overlapping semantics between negative samples which should be semantic-distinct, e.g. two medical images sharing the same disease are contrasted which brings noise~\cite{emnlp2022_MedCLIP}. Once directly learning, cross-modal representations of the same disease are falsely pulled apart, making the model unable to capture the disease-corresponding image feature. \textbf{(b) Semantic Shifting Problem:} Radiologists have writing preferences, e.g. biased for their own familiar concepts and observation view towards similar
\begin{figure}
	\centering
	\includegraphics[width=0.9\textwidth]{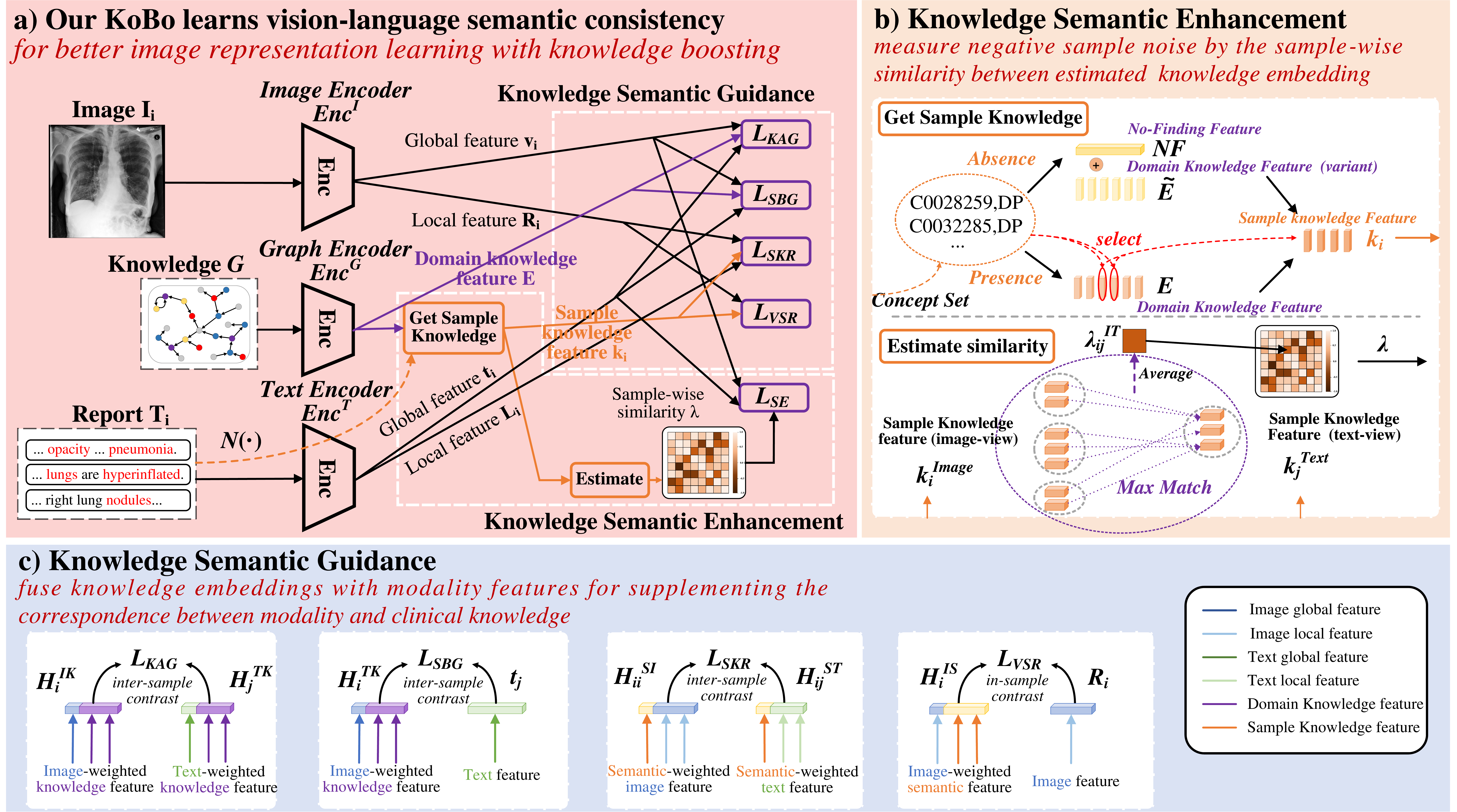}
	\caption{Overview of our proposed architecture, where additional clinical knowledge is embedded in. Image encoder, text encoder, graph encoder, knowledge semantic enhancement module, and knowledge semantic guidance module are presented.} 
	\label{framework}
\end{figure}
visual features, and inclined for negation expression towards opposite visual features. Distinct concepts describing the same image are morphologically dissimilar for text encoder, while the negation expression of concepts is morphologically similar~\cite{negbio}. Once lack of concept correlation and negation identification, representations with similar semantics are falsely pushed apart and those with opposite semantics are falsely pushed together, interfering with the learning of significant representation\cite{he2022learning}.

Rethinking the existing methods and challenges of medical contrastive vision-language pre-training~\cite{iccv2021Gloria,pmlr2022ConVIRT,nips2022_MGCA,emnlp2022_MedCLIP,miccai2022_RGCT}, the lack of clinical knowledge constraints in dual-free-encoding contrastive learning structure is the key problem. Existing methods utilize sample-wise differences to learn mutual information between modalities, improving the representation quality based on the correspondence of learned mutual information and clinical knowledge. However, semantic overlap reduces the learning efficiency of mutual information with the noisy difference, and the mentioned correspondence is vulnerable to semantic shifting. Therefore, if we are able to embed an unbiased, comprehensive representation as knowledge boosting, it will reduce the negative noise and supplement the lacking correspondence. It motivates us to measure the noise with similarities between knowledge representation, and fuse the correspondence between knowledge and modality.

In this paper, we propose a novel knowledge-boosting medical contrastive vision-language pre-training framework (KoBo). Our contributions are as followed. \textbf{1)} Our KoBo pre-trains a powerful image encoder including visual information corresponding with the disease described in texts, where knowledge is embedded in our paradigm (Fig.\ref{paradigm}) to boost the learning of vision-language consistency. \textbf{2)} We propose Knowledge Semantic Enhancement (KSE) module to reduce the negative sample noise with the similarity between open-set sample-wise knowledge embeddings. \textbf{3)} We propose Knowledge Semantic Guidance (KSG) module to adjust the semantic shifting during pre-training, fusing the modality feature with unbiased knowledge embeddings for supplementing the correspondence between modality mutual information and clinical knowledge.
	
\section{Methodology}
Our Knowledge-Boosting Contrastive Vision-Language Pre-training framework (Fig.\ref{framework}) boosts vision-language learning with additional clinical knowledge. It contains two modules: KSE for reducing the negative effect of semantic overlap, and KSG for adjusting semantic shifting, aimed at learning effective representation by maximizing semantic consistency between paired image and text features.
	
\subsection{Framework Formulation}
In the framework, a powerful image encoder $Enc^{I}$ and text encoder $Enc^{T}$ is pre-trained, alongside a graph encoder $Enc^{G}$. Given a pair of medical image and diagnostic report $\{I_{i}, T_{i}^{Report}\}, I_{i} \in \mathbb{R}^{H \times W \times C}$, a sentence $T_{i}^{Sent}$ is randomly selected from $T_{i}^{Report}$ as a caption comprised of several tokens $\{w_{1}, w_{2},...,w_{N_{L}}\}$. $Enc^{I}$ outputs global feature $z_{i}^{I,G}$ and local feature $z_{i}^{I,L}$ for $N_{I}$ sub-regions, which is from the intermediate feature map. $T_{i}^{Sent}$ is fed into $Enc^{T}$, obtaining global sentence feature $z_{i}^{T,G}$, and local token feature $z_{i}^{T,L}$. Distinct projectors are applied to map features into embeddings with lower semantic dim $D_{S}$, finally getting global and local image embeddings $v_{i} \in \mathbb{R}^{D_{S}}, R_{i}=\{r_{i1},r_{i2},...,r_{iN_{I}}\} \in \mathbb{R}^{N_{I} \times D_{S}}$, and text embedding $t_{i} \in \mathbb{R}^{D_{S}}, L_{i}=\{l_{i1},l_{i2},...,l_{iN_{L}}\} \in \mathbb{R}^{N_{L} \times D_{S}}$.
	
Besides using reports and images as the input for our pre-training network, we also input an external knowledge graph to the whole framework for improving the correspondence of modality features and clinical knowledge. The knowledge refers to relations between clinical pathology concepts in the radiology domain in the format of triplet $\mathcal{G}=\{(c_{h_{k}},\emph{r}_{k},c_{t_{k}})\}^{N_{G}}_{k=1}$, such as UMLS~\cite{UMLS_KG}. Domain knowledge embedding for each concept $E=\{e_{s}\}^{N_{E}}_{s=1} \in \mathbb{R}^{N_{E} \times D_{S}}$ is the output of $Enc^{G}(\mathcal{G})$.
	
\subsection{Knowledge Semantic Enhancement}
\label{Section 2.2}
To relieve the semantic overlap problem, where negative sample noise harms the effective learning of vision-language mutual information, we propose a semantic enhancement module to identify the noise using sample-wise similarities. The similarity is estimated upon sample knowledge $k_{i}$, calculated from domain knowledge embedding $E$ and concept set from texts with negation marker.
\\
\textbf{Getting Sample knowledge}: Firstly, we acquire a concept set that contains pathology concepts extracted from texts with Negbio $\mathcal{N}(\cdot)$~\cite{negbio}. The image-view concept set which involves the overall observation is from the whole report, while the text-view set only covers the chosen sentence. Secondly, the image and text sample knowledge, as an auxiliary semantic estimation, is selected from domain knowledge embedding $E$ according to the corresponding concept set from the report and sentence respectively, if not considering the negation problem.

Furthermore, considering the challenge that negation expression of concepts commonly exists in radiology reports, which has opposite semantics with similar morphology for text encoder (converging shifting), we randomly generate a \emph{No Finding} embedding $\mathcal{NF}$ and a variant of domain knowledge embedding $\widetilde{E} = \{\widetilde{e}_{1}, \widetilde{e}_{2}, ..., \widetilde{e}_{N_{E}}\}$ of the same size as $E$ with Xavier distribution. Upon the negation mark of concept, sample knowledge embedding $k_{i}=\{k_{i,s}\}^{N_{ES}}_{s=1}$ is denoted below:
\begin{equation}
	k_{i,s} = \begin{cases}
		e_{i,s} & c_{i,s} \in \mathcal{N}(T_{i}), P(c_{i,s}) \neq Neg \\ 
		\epsilon \cdot \mathcal{NF} +(1-\epsilon) \widetilde{e}_{i,s} & c_{i,s} \in \mathcal{N}(T_{i}), P(c_{i,s}) = Neg
	\end{cases}
\end{equation}
where $P$ is the negation mark of concepts, and $e_{i,s}, \widetilde{e}_{i,s}$ is the corresponding position of $c_{i,s}$ in $E$ and $\widetilde{E}$. $\epsilon$ tunes the variance of negative sample knowledge. $k^{Image}_{i,s}$ and $k^{Text}_{i,s}$ are $k_{i}$ from the image-view and text-view concept set.
\\
\textbf{Estimation of Similarities:} The semantic similarity is calculated upon sample knowledge. For each image-text pair, a max-match strategy is adopted to match each two sample knowledge embedding with the most similar one for calculating cosine similarities. Sample-wise similarities are aggregated with averages.
\begin{equation}
	\lambda_{ij}^{IT} = \frac{1}{N_{ES'}}\sum^{N_{ES'}}_{s=1}\max^{N_{ES}}_{s'=1}(k_{i,s}^{Image})^{T}k_{j,s'}^{Text},
	\lambda_{ij}^{TI} = \frac{1}{N_{ES}}\sum^{N_{ES}}_{s=1}\max^{N_{ES'}}_{s'=1}(k_{i,s}^{Text})^{T}k_{j,s'}^{Image}
\end{equation}
where $N_{ES}$ is the number of concepts in $T_{i}^{Sent}$, while $N_{ES'}$ is that in $T_{i}^{Report}$.
\\
\textbf{Knowledge Semantic Enhancement Loss}: We utilize the sample-wise semantic similarity to estimate negative sample noise, placed in the sample weight of the contrastive loss~\cite{pmlr2022ConVIRT,pmlr2021_CLIP}, where paired cross-modal embedding are pushed together and unpaired ones are pulled apart. The importance of estimated noisy negative samples is relatively smaller for a subtle pulling between cross-modal embeddings. The semantic enhancement loss is below:
\begin{equation}
	\label{eq.1}
	\mathcal{L}_{SE} = -\frac{1}{N} \sum_{i=1}^{N}(\log \frac{ \exp(v_{i}^{T}t_{i}/\tau_{G})}{\sum\limits_{j=1}\limits^{N} (1-\lambda^{IT}_{ij}) \exp(v_{i}^{T}t_{j}/\tau_{G})} + \log \frac{ \exp(t_{i}^{T}v_{i}/\tau_{G})}{\sum\limits_{j=1}\limits^{N} (1-\lambda^{TI}_{ij}) \exp(t_{i}^{T}v_{j}/\tau_{G})})
\end{equation}
where $\tau_{G}$ is the global temperature, and $\lambda^{IT}$, $\lambda^{TI}$ is the sample similarity measurement. specifically, $\lambda_{i,i}$ is fixed to zero to persist the positive sample weight.

\subsection{Knowledge Semantic Guidance}
In this section, we propose a semantic guidance module to solve the semantic shifting problem. Utilizing sample knowledge from Section~\ref{Section 2.2} which contains concept correlation and negation information, the adverse effects of both disperse and converging shifting are alleviated by fusing domain-sample knowledge with global-local modality embeddings. We design four contrast schemes: knowledge anchor guidance for adjusting disperse shifting, semantic knowledge refinement for filtering converging shifting, vision semantic response for consolidating knowledge fusion, and semantic bridge guidance for narrowing the modality gap.
\\
\textbf{Knowledge Anchor Guidance}: Disperse shifting will be adjusted if there are unbiased anchors in semantic space as priors to attract modality embeddings towards clinical semantics, and domain knowledge embedding does a good job. We define knowledge fused embeddings $H_{i}^{IK} = ATTN(v_{i}, E, E)$ and $H_{i}^{TK} = ATTN(t_{i}, E, E)$, and  $ATTN(Q,K,V)$ means the attention function~\cite{iccv2021Gloria}:
\begin{equation}
\mathcal{L}_{KAG} = -\frac{1}{N} \sum_{i=1}^{N}(\log \frac{ exp(H_{i}^{IK}\cdot H_{i}^{TK}/\tau_{G})}{\sum_{j=1}^{N} exp(H_{i}^{IK}\cdot H_{j}^{TK}/\tau_{G})} + \log \frac{ exp(H_{i}^{TK} \cdot H_{i}^{IK}/\tau_{G})}{\sum_{j=1}^{N} exp(H_{i}^{TK}\cdot H_{j}^{IK}/\tau_{G})})
\end{equation}
where image-weighted and text-weighted knowledge is globally contrasted.
\\
\textbf{Semantic Knowledge Refinement}: Wrong-converging pairs have distinct intrinsic responses on sample knowledge from image and text. Hence, we propose to utilize sample knowledge to refine these falsely gathered dissimilar pairs. We define $H_{ij}^{SI} = ATTN(k^{Text}_{i}, R_{j}, R_{j})$ and $H_{ij}^{ST} = ATTN(k^{Text}_{i}, L_{j}, L_{j})$:
\begin{equation}
\mathcal{L}_{SKR} = -\frac{1}{N}\sum^{N}_{i=1}\log\frac{exp(\frac{1}{N_{ES} \cdot \tau_{L}}\sum^{N_{ES}}_{k=1}H_{iik}^{SI}\cdot H_{iik}^{ST})}{\sum^{N}_{j=1}exp(\frac{1}{N_{ES} \cdot \tau_{L}}\sum^{N_{ES}}_{k=1}H_{ijk}^{SI}\cdot H_{ijk}^{ST})}
\end{equation}
where local semantic-weighted image and text embeddings are contrasted.
\\
\textbf{Vision Semantic Response}: Instead of matching single token with image sub-regions in~\cite{iccv2021Gloria}, we propose to match the concept with sub-regions. As the concept is a more complete and atomic semantic unit, local response upon concept will better guide the representation learning with a fine-grained semantic match through an in-sample contrast. We define $H_{i}^{IS} = ATTN(R_{i}, k^{Text}_{i}, k^{Text}_{i})$, and the fusion of knowledge will be consolidated as below:
\begin{equation}
\mathcal{L}_{VSR} = -\frac{1}{N\cdot N_{I}}\sum^{N}_{i=1}\sum^{N_{I}}_{k=1}\log\frac{exp(H_{ik}^{IS}\cdot r_{ik}/\tau_{L})}{\sum^{N_{I}}_{k'=1}exp(H_{ik}^{IS}\cdot r_{ik'}/\tau_{L})}
\end{equation}
where there is an in-sample local contrast between $H_{i}^{IS}$ and vision features.
\\
\textbf{Semantic Bridge Guidance}: We propose to narrow disperse shifting enlarged by the modality gap between vision and language. Specifically, the gap is bridged by the fusion of domain knowledge which is better compatible with text:
\begin{equation}
\mathcal{L}_{SBG} = -\frac{1}{N}\sum^{N}_{i=1}(\log\frac{exp(H_{i}^{IK}\cdot t_{i}/\tau_{G})}{\sum^{N}_{j=1}exp(H_{i}^{IK}\cdot t_{j}/\tau_{G})}+\log\frac{exp(t_{i} \cdot H_{i}^{IK}/\tau_{G})}{\sum^{N}_{j=1}exp(t_{i} \cdot H_{j}^{IK}/\tau_{G})})
\end{equation}
where the image-weighted domain knowledge is contrasted with text features between samples. Finally, $\mathcal{L}_{SG}$ is aggregated by these four parts as below:
\begin{equation}
\mathcal{L}_{SG} = \lambda_{1} \mathcal{L}_{KAG} + \lambda_{2} \mathcal{L}_{SKR} + \lambda_{3} \mathcal{L}_{VSR} + \lambda_{4} \mathcal{L}_{SBG}
\end{equation}

\section{Experiment}
\begin{table}[t]
\centering
\caption{Comparison results in eight downstream tasks. (*) defines that official pre-trained weight is used, and the remaining methods are reproduced with the same batch size, pre-processing and the evaluation. CLS, RR, SR, and SEG mean classification, retrieval, semantic relatedness and semantic segmentation, V or L means vision and language tasks. Few-shot-Frozen means the frozen encoder of the backbone and only 1\% of total training data. ResNet-50 is the equal-comparing backbone except for KoBo-Vit. The best two results are highlighted in underlined red and violet.}\label{comparison}
\resizebox{\textwidth}{!}{
	\begin{tabular}{c|cccccccc}
		\toprule
		\multirow{5}{*}{Method} & \multicolumn{5}{c|}{\textbf{Zero-shot}}                                                                                                                                                                                                                                                                                                                     & \multicolumn{3}{c}{\textbf{Few-shot-Frozen}}                                                                                                                                                      \\ \cline{2-9} 
		& \begin{tabular}[c]{@{}c@{}}\textbf{CLS(V+L)}\\ CheXpert\\ (Auroc)\end{tabular} & \begin{tabular}[c]{@{}c@{}}\textbf{RR(V)}\\ CheXpert5X200\\ (mAP)\end{tabular} & \begin{tabular}[c]{@{}c@{}}\textbf{RR(V+L)}\\ MIMIC\\ (mAP)\end{tabular} & \begin{tabular}[c]{@{}c@{}}\textbf{SR(L)}\\ UMNSRS\\ (Pearson)\end{tabular} & \multicolumn{1}{c|}{\begin{tabular}[c]{@{}c@{}}\textbf{SR(L)}\\ MIMIC\\ (Pearson)\end{tabular}} & \begin{tabular}[c]{@{}c@{}}\textbf{CLS(V)}\\ CheXpert\\ (Auroc)\end{tabular} & \begin{tabular}[c]{@{}c@{}}\textbf{SEG(V)}\\ SIIM\\ (Dice)\end{tabular} & \begin{tabular}[c]{@{}c@{}}\textbf{CLS(V)}\\ Covidx\\ (Acc)\end{tabular} \\ \hline
		CLIP\cite{pmlr2021_CLIP}(*)                          & 0.4702                                                           & 0.2544                                                                & 0.7577                                                     & 0.1985                                                          & \multicolumn{1}{c|}{-0.2879}                                                        & 0.5748                                                           & /                                                               & 0.8975                                                       \\
		ConVIRT\cite{pmlr2022ConVIRT}                          & 0.8252                                                           & 0.3808                                                                & \color{red}{\textbf{\underline{0.8482}}}                                            & \color{violet}{\textbf{0.2506}}                                                 & \multicolumn{1}{c|}{0.1429}                                                         & 0.8548                                                           & 0.4992                                                          & 0.9475                                                       \\
		Gloria\cite{iccv2021Gloria}                           & 0.8257                                                           & 0.3875                                                                & 0.8390                                                     & 0.2294                                                          & \multicolumn{1}{c|}{0.1100}                                                         & 0.8492                                                           & 0.5479                                                          & 0.9250                                                       \\
		MGCA\cite{nips2022_MGCA}                             & 0.8496                                                           & 0.3906                                                                & 0.8428                                                     & 0.1889                                                          & \multicolumn{1}{c|}{0.1809}                                                         & 0.8616                                                           & 0.5696                                                          & 0.9375                                                       \\
		MedCLIP\cite{emnlp2022_MedCLIP}(*)                       & 0.7805                                                           & \color{red}{\textbf{\underline{0.4298}}}                                                       & 0.7258                                                     & 0.2032                                                          & \multicolumn{1}{c|}{-0.1321}                                                        & 0.8214                                                           & 0.5619                                                          & 0.9325                                                       \\ \hline
		\textbf{KoBo}            & \color{violet}{\textbf{0.8590}}                                                  & 0.3918                                                                & \color{violet}{\textbf{0.8467}}                                            & \color{red}{\textbf{\underline{0.2563}}}                                                 & \multicolumn{1}{c|}{\color{violet}{\textbf{0.3712}}}                                                & \color{violet}{\textbf{0.8628}}                                                  & \color{violet}{\textbf{0.6393}}                                                 & \color{red}{\textbf{\underline{0.9550}}}                                              \\
		\textbf{KoBo-Vit}        & \color{red}{\textbf{\underline{0.8635}}}                                                  &\color{violet}{\textbf{0.4123}}                                                       & 0.8455                                                     & 0.1824                                                          & \multicolumn{1}{c|}{\color{red}{\textbf{\underline{0.4229}}}}                                                & \color{red}{\textbf{\underline{0.8660}}}                                                  &\color{red}{\textbf{\underline{0.6554}}}                                             & \color{violet}{\textbf{0.9525}}                                       \\ \bottomrule
	\end{tabular}
}
\end{table}
\begin{figure}
\centering
\includegraphics[width=\textwidth]{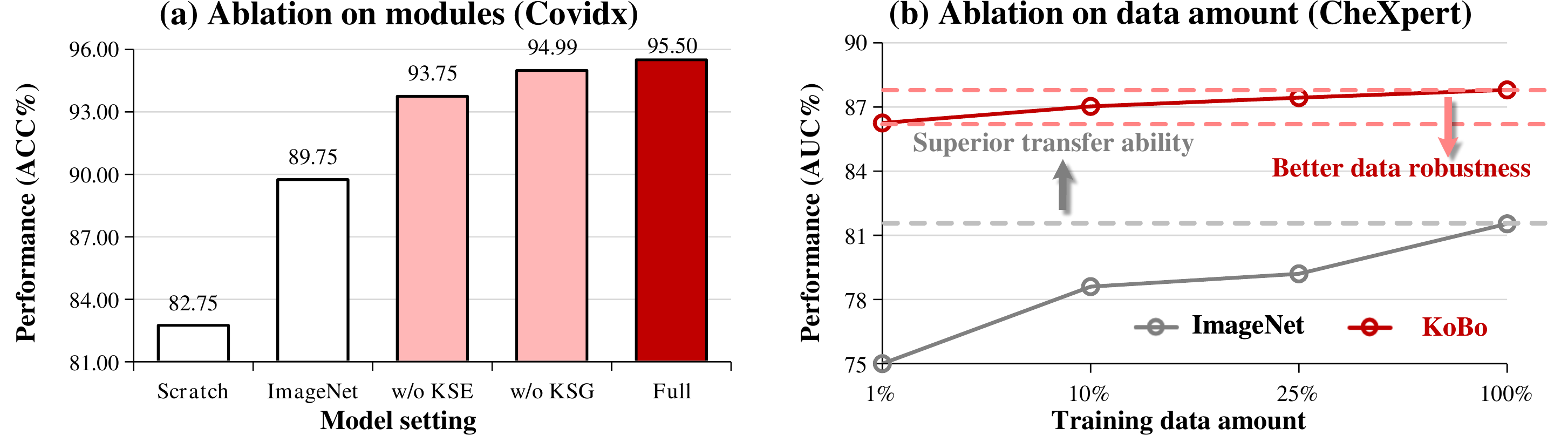}
\caption{\textbf{(a)} Module ablation study of our KoBo framework is performed on Covidx dataset compared with ImageNet and random initialization, upon few-shot frozen setting. \textbf{(b)} Data ablation study is performed on CheXpert with frozen setting when classification training data amount reduces to 25\%, 10\% and 1\%.}
\label{abalation}
\end{figure}
\textbf{Experiment Protocol:} Pre-training performs on MIMIC-CXR~\cite{MIMIC-CXR-JPG} following the pre-process style of \cite{RATCHET}. The impression section of reports and frontal view of images are selected to generate 203k image-report pair. Five downstream task datasets (CheXpert~\cite{CheXpert}, Covidx~\cite{Covidx}, MIMIC-CXR, UMNSRS~\cite{UMNSRS} and SIIM~\cite{SIIM-ACR}) are applied on eight tasks. Semantic relatedness is to verify the text understanding of radiology concepts, where text embedding with certain prompts predicts the relatedness. A new semantic relatedness benchmark is generated from MIMIC-CXR, adding in the extra negation discriminating. CheXpert5X200~\cite{iccv2021Gloria}(Multi-classification) is from CheXpert, and CheXpert-labeller\cite{CheXpert} generates retrieval labels in MIMIC-CXR. More details are in appendix.
\\
For implementation, ResNet50~\cite{resnet} and Vit~\cite{vit} are image encoder, and BioClinicalBERT~\cite{bioclinicalBERT} is the text encoder. CompGCN with LTE~\cite{www2022_CompRGCN_LTE} is our graph encoder, and domain knowledge contains 10,244 concepts in UMLS which exist in MIMIC-CXR. Negbio~\cite{negbio} combined with UMLS disambiguation tool~\cite{UMLS_KG} serves as $\mathcal{N}(\cdot)$. Embeddings are projected into the dim of 256. Pre-training has the batch size of 100 and max epochs of 50 based on Pytorch on two RTX3090 GPUs.
\begin{figure}
	\centering
	\includegraphics[width=\textwidth]{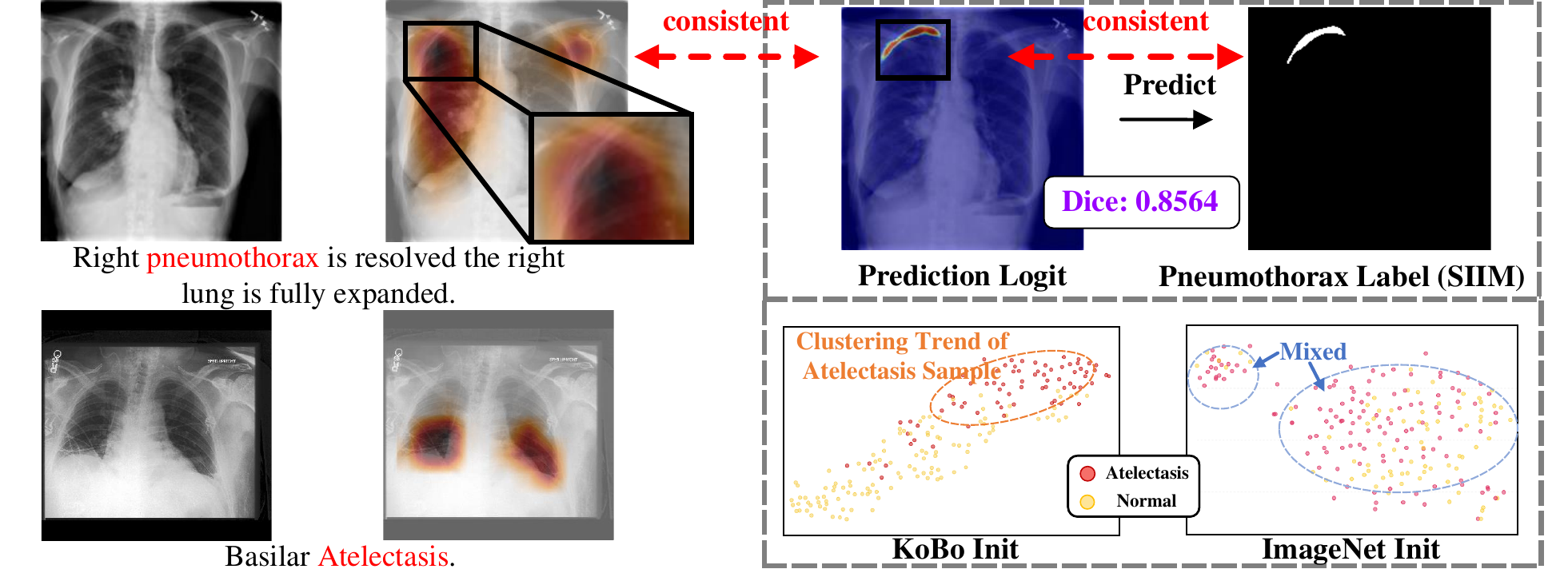}
	\caption{Visualization of pneumothorax and atelectasis. AblationCAM~\cite{ablationCAM} generates the class activate map (CAM) upon last layer of Kobo-ResNet. There is strong consistency between CAM, prediction logits and segmentation label. t-SNE~\cite{nips2022_MGCA} is applied on image embedding from CheXpert-valid, showing gathering cluster trend of disease samples.}
	\label{visual}
\end{figure}
Adam optimizer with the learning rate of 5e-5 and ReduceLR scheduler are applied. $\tau_{G}$ is 0.07 and $\tau_{L}$ is 0.1. $\lambda$ in KSG loss are all 0.25, while $\epsilon$ in KSE loss is 0.1.
\\
\\
\textbf{Comparison Study:} Table~\ref{comparison} verifies our powerful representation ability, reaching state-of-art in classification, segmentation, and semantic relatedness compared with existing vision-language pre-training tasks, while our method is also top two for retrieval. In zero-shot classification tasks, our KoBo outperforms MGCA and ConVIRT 0.94\% and 3.38\% respectively, exceeding most methods even in their training setting. For CheXpert5X200, our framework is second only to MedCLIP which presents a superior performance in this dataset. In three few-shot setting task, our KoBo has an absolute leading position.
\\
\\
\textbf{Ablation Study:} As is demonstrated in Fig.\ref{abalation}, we perform module ablation and data amount ablation. \textbf{(a)} For module ablation, both modules bring benefits in representation learning and are respectively effective. When KSG module is removed, our KoBo also extracts effective feature related to pneumonia with a subtle decrease of 0.51\%. When KSE is removed, there is a reduction of 1.25\% accuracy. \textbf{(b)} For data amount ablation, KoBo has better data robustness with a subtle decrease when training data reduce to 1\%. KoBo also has a superior transfer ability with an absolutely better AUC with 1\% data than ImageNet with all training data than ImageNet with all training data.
\\
\\
\textbf{Qualitative Analysis:} In Fig.\ref{visual}, our Kobo has learned fine-grained and effective image feature with the fusion of knowledge modeling. The deepest region in the first image gathered on the top left side, showing an obvious expansion on the right lung. There is consistency with the expert annotation and our output logit. The precise location of atelectasis region in CAM of second image and clustering trend interpret for the increase in zero-shot classification.

\section{Conclusion}
In our paper, we propose a Knowledge-Boosting Contrastive Vision-Language Pre-traing framework (KoBo). Sample and domain knowledge are used to differentiate noisy negative samples and supplement the correspondence between modality and clinical knowledge. Our experiments on eight tasks verify the effectiveness of our framework. We hope that our work will encourage more research on knowledge-granularity alignment in medical vision-language learning.

\subsubsection{Acknowledgements} This work was supported in part by the National Natural Science Foundation under grants (62171125, 61828101), CAAI-Huawei Mind Spore Open Fund, CANN (Compute Architecture for Neural Networks), Ascend AI Processor, and Big Data Computing Center of Southeast University.

\bibliographystyle{splncs04}
\bibliography{references}
\end{document}